\documentclass[10pt,twocolumn,letterpaper]{article}

\usepackage{iccv}
\usepackage{times}
\usepackage{epsfig}
\usepackage{graphicx}
\usepackage{amsmath}
\usepackage{amssymb}

\usepackage[pagebackref=true,breaklinks=true,letterpaper=true,colorlinks,bookmarks=false]{hyperref}
\usepackage{cleveref}

\iccvfinalcopy %

\def\iccvPaperID{****} %

\ificcvfinal\fi

\begin{document}

\title{Training CLIP models on Data from Scientific Papers}

\author{Calvin Metzger\\
TU Wien\\
{\tt\small calvin.metzger@student.tuwien.ac.at}
}

\maketitle
\ificcvfinal\thispagestyle{empty}\fi

\begin{abstract}
Contrastive Language-Image Pretraining (CLIP) models are able to capture the semantic relationship of images and texts and have enabled a wide range of  applications, from image retrieval to classification. These models are trained with datasets extracted from web crawls, which are of large quantity but limited quality. This paper explores whether limited amounts higher quality data in a specific domain improve the general performance of CLIP models. To this purpose, we extract text-image data from scientific papers hosted in the arXiv and PubMed Central repositories. Experiments on small-scale CLIP models (ViT B/32) show that model performance increases on average, but only moderately. This result indicates that using the data sources considered in the paper to train large-scale CLIP models is a worthwile research direction.

\end{abstract}

\section{Introduction}

In recent years, Contrastive Language-Image Pretraining (CLIP) models \cite{DBLP:conf/icml/RadfordKHRGASAM21} have emerged as a transformative advancement in the field of computer vision and natural language processing. These models excel at understanding the intricate relationship between textual descriptions and visual content, enabling a wide range of applications, from image retrieval to zero-shot classification. However, the effectiveness of CLIP models hinges critically on the quality and diversity of their training data.

CLIP models are commonly trained using images and associated alt-text extracted from large scale web crawl archives \cite{commoncrawl}. This data provides the quantity needed to successfully train such a large model. However, the quality of this data is generally quite poor. A usual strategy against this is to filter the data, e.g. according to the similarity of images and texts of previous CLIP models.

Recent research in the context of NLP suggests that mixing a limited amount of data of higher quality into the training dataset improves general model performance \cite{xie2023doremi,madaan2022language}, even if this data only covers a limited range of domains. Hence, this paper explores this potential for CLIP models by extending the dataset with image-text pairs extracted from scientific papers, which are assumed to be of high quality.

Two sources of scientific papers are considered. Firstly, the arXiv repository, which hosts a large amount of papers in a variety of - mostly quantitative - fields. Secondly, the PubMed Central (PMC) repository, which provides open access to papers in the biomedical domain.

The CLIP models trained on this dataset are evaluated using the evaluation suite introduced by Gadre et al. \cite{DBLP:journals/corr/abs-2304-14108}, which includes the most commonly used datasets and tasks to evaluate CLIP models. Since the domain distribution of the evaluation datasets is quite different from the domain distribution in arXiv and PMC, this paper investigates whether a training a model on our dataset improves performance in spite of the limited domain coverage.

For reproducibility, the code, data and models are open sourced at \url{https://github.com/nopperl/clip_arxiv_pmc}.

\section{Related works}
A CLIP model pretrained on image-text pairs extracted from PMC is BioMedClip \cite{DBLP:journals/corr/abs-2303-00915}, which showed promising results. Similarly, PMC-CLIP \cite{lin2023pmcclip} is trained on a filtered subset of image-text pairs extracted from PMC. To our knowledge, no CLIP model trained on image-text pairs from arXiv exist. However, there have been efforts to extract such data for similar tasks \cite{DBLP:conf/jcdl/SiegelLPA18}.

\section{Data collection}

The dataset used to train the CLIP model consists of three subsets:

\begin{itemize}
	\item CommonPool: images and alt-texts extracted from Common Crawl \cite{commoncrawl}
	\item arXiv: figures and captions extracted from papers hosted on the arXiv repository \cite{arxiv}, and
	\item PMC: figures and captions extracted from papers part of the PubMed Central Open Access Subset \cite{pmcoas}.
\end{itemize}

\Cref{tab:data.stats} provides summary statistics for the collected subsets. Note that the CommonPool data is retrieved using the workflow provided by Gadre et al. \cite{DBLP:journals/corr/abs-2304-14108}. Due to limitations in storage space and computational resources, only the small scale of their dataset is used.

\begin{table}
	\begin{tabular}{l r r}
		Dataset & \# figures & avg caption length \\
		CommonPool & 11778443 & 45.21 \\
		arXiv & 1117377 & 266.96 \\
		PMC & 766855 & 464.91 \\
	\end{tabular}
	\caption{Summary statistics of the collected datasets.}
	\label{tab:data.stats}
\end{table}

An underlying assumption of our work is that data from arXiv and PMC is of higher quality than CommonPool. A rough proxy measure for that is the caption length, which is significantly higher for the arXiv and PMC datasets than the CommonPool caption length. This suggests (but is insufficient to prove) that captions of datasets collected in this paper are more detailed and of higher quality.

\Cref{subsec:data.arxiv,subsec:data.pmc} describe the data collection workflow for the arXiv and the PMC subset, respectively. \Cref{subsec:data.decontamination} describes steps taken to decontaminate the collected dataset.

\subsection{arXiv}
\label{subsec:data.arxiv}

The arXiv repository provides papers both in printable format (i.e., as PDF files) and in their source format. While the PDF files can easily be downloaded in bulk from the arXiv dataset hosted on Google Cloud, the source files are hosted on a requester pays Amazon S3 bucket \footnote{\url{https://info.arxiv.org/help/bulk_data_s3.html}}. It is possible to use the PDF files directly to extract figures and captions using extraction pipelines based on machine learning and OCR \cite{DBLP:conf/jcdl/SiegelLPA18}. However, these pipelines inherently introduce failure modes due to the possibility of inaccurate outputs of their models. On the other hand, using structured source files allows to extract figures and captions in an accurate way. Since the printable versions of papers hosted on arXiv are compiled by arXiv based on the sources submitted by authors, the sources need to adhere to a specific format mandated by arXiv. This guarantee of a specific format helps with the accurate extraction of figures and captions. Hence, it is decided to base the extraction pipeline on paper source files.

The source files of all papers hosted on arXiv up to (and including) 2020-12-31 are downloaded from their requester pays Amazon S3 bucket. The data makes up 1.4TB and is divided into tar archives in chronological order with a size of roughly 500M each. Each tar archive contains gzipped source files or, if not available, the PDF version of papers in chronological order.

When extracted using gzip, the paper source files are of a variety of different formats, including Ghostscript, HTML, single \TeX files or tar archives containing a \LaTeX project. The former three formats are used by only a small proportion of papers and - due to their nature of being single files - do not contain images other than simple vector graphics such as graphs, if at all. Hence, only tar archives of \LaTeX projects are considered for the dataset.

For this work, the relevant files of a tar archive are \LaTeX and \TeX files with the extension \texttt{.tex} and image files. Images used in \LaTeX projects are stored using the (Encapsulated) PostScript format with the extensions \texttt{.ps} or \texttt{.eps}. Furthermore, since arXiv uses pdfLaTeX to compile the sources, images can also be stored using the JPEG, PNG, GIF and PDF formats. Hence, only image files with one of the extensions used by pdfLaTeX or \LaTeX are considered for extraction: \texttt{.jpg, .jpeg, .gif, .png, .pdf, .eps, .ps}.

The extraction pipeline iteratively processes each .tex file. The .tex files are consistently encoded using the \texttt{ISO-8859-1}. The \TeX source is decoded and then parsed using the TexSoup package \cite{TexSoup} to query for all \verb|\includegraphics| commands. Note that all \verb|\newcommand| definitions are ignored, which means that \verb|\includegraphics| aliases are not detected. The \verb|\includegraphics| command indicates that the image at the specified path is to be included in the document. Since the compilation by arXiv is executed from the root directory of the archive, the path is guaranteed to be relative to this root directory. If the path corresponds to an image file available in the tar archive, the \verb|\includegraphics| has a neighboring \verb|\caption| command and there is no other neighboring \verb|\includegraphics| command, the graphic and caption are added to the dataset. This workflow is agnostic to the environment used to contain the graphic and its caption.

An important case are figures which consist of multiple graphics. These are complicated to handle, since it is unclear which part of the caption semantically corresponds to which graphic. In general, the restriction of considering only \verb|\includegraphics| commands that have no neighboring \verb|\includegraphics| commands is introduced to ignore these kinds of figures. However, subfigures as a structured subcase of multiple-graphics figures are handled in a rudimentary way in the workflow explained above. Since the workflow is agnostic to the environment, a graphic in the \texttt{subfigure} environment will be included with its neighboring \texttt{caption}, but not with the \texttt{caption} of the parent \texttt{figure}.

At this stage, the caption consists of \LaTeX source text, which is considerably different from the plain text used in CommonPool. To bring the caption into consistent format with CommonPool, the pylatexenc package \cite{pylatexenc} is used to convert the caption text into unicode plaintext. Note that this only handles references and citations by replacing them with \verb|<ref>| and \verb|<cit.>|, respectively.

Finally, the images are converted to be consistent with the CommonPool dataset. That is, each image is converted to RGB and \texttt{.jpg} using the Pillow package \cite{Pillow}. Furthermore, all images are resized to 512px.

\subsection{PubMed Central Open Access Subset}
\label{subsec:data.pmc}

The PubMed Central biomedical paper repository provides a subset of openly-licensed papers. Similar to arXiv, these are distributed as PDF files. However, XML files for semantic markup are also available, which include information about figures. Similar to the arXiv dataset, the pipeline is based on these files in order to guarantee accurate extraction.

All papers up to 2023-08-20 are downloaded from the publicly available FTP server at \url{https://www.ncbi.nlm.nih.gov/pmc/tools/ftp/}. The papers are collected in a two-level directory hierarchy. Each paper corresponds to a gzipped tar archive. The archive consists of a single .nxml file containing the markup of the paper. Images in the paper are stored using .jpg and .gif files. Since the resolution of .gif files is only thumbnail sized and way too small for the target resolution of 512px, only .jpg files are considered.

The nxml file is parsed using the lxml package \cite{lxml} and queried for the \texttt{fig} element. Only english captions are considered, so figures which have a \texttt{lang} attribute set that does not start with \texttt{en} are ignored. Similar to the arXiv subset, if the path of the \texttt{graphic} element in the figure does not correspond to an existing .jpg file in the tar archive, it is ignored. Otherwise, the graphic and caption are added to the dataset.

Due to resource limitations, only 10 out of 257 top-level directories are processed, leading to only a fraction of available figures being extracted.

\subsection{Decontamination}
\label{subsec:data.decontamination}
Since the dataset was assembled from a wide range of papers, it might contain images that are also present in the evaluation datasets. In order to properly evaluate models trained using the dataset, it is important to prevent evaluation set leakage and to remove those images. Hence, the collected arXiv and PMC datasets are decontaminated against the datasets contained in the evaluation suite introduced by Gadre et al. \cite{DBLP:journals/corr/abs-2304-14108}, which covers most commonly used datasets used to evaluate CLIP models.

Following Gadre et al. \cite{DBLP:journals/corr/abs-2304-14108}, the similarity of images in the dataset to evaluation dataset images is measured using the model introduced by Yokoo \cite{DBLP:journals/corr/abs-2112-04323}. Samples with a similarity score higher than 0.604169 are removed from the dataset. This filter removes 0.7\% and 0.4\% of the samples in the arXiv and PMC subset, respectively.

\paragraph{Further filtering}

Note that, since the data is sourced from research papers intended for open publication, the amount of NSFW or toxic data is assumed to be low. Furthermore, filtering based on machine learning models such as Detoxify \cite{Detoxify} harbors the risk of introducing biases to the data due to false positives. Hence, since the advantage is assumed to be low compared to potential disadvantages, no further filtering workflow is applied to the collected data.

\section{Experiments}

To ascertain whether the data collected in this paper improves the performance of CLIP models, the same model architecture is trained and evaluated with and without the additional data. The small scale model architecture introduced by Gadre et al. \cite{DBLP:journals/corr/abs-2304-14108} based on OpenCLIP is used for this purpose. The visual encoder is a ViT-B/32. The models are trained with a learning rate of 5e-4 with 500 warmup steps, a batch size of 4096 and the AdamW optimizer with $\beta_2 =0.98$. For more details about the model, refer to Gadre et al. \cite{DBLP:journals/corr/abs-2304-14108}.

Similar to the model architecture, the baseline CommonPool dataset is also taken from Gadre et al. \cite{DBLP:journals/corr/abs-2304-14108}. Note that the small scale, i.e. a random subset of roughly 12M observations is used.

To get results for our collected data, two models are trained. One model is trained solely using the collected dataset, while the other is trained both the CommonPool and our dataset. The models are trained using the same amount of steps as the baseline (12.8M), with observations in the training data sampled uniformly without replacement per step. This leads to the data proportions for the latter model listed in \Cref{tab:data.weights}.

\begin{table}
	\centering
	\begin{tabular}{l r}
		Subset & Proportion \\
		CommonPool & 86\% \\
		arXiv & 8\% \\
		PMC & 6\%
	\end{tabular}
	\caption{Data subset proportions for the model trained on CommonPool, arXiv, and PMC.}
	\label{tab:data.weights}
\end{table}

Following Gadre et al. \cite{DBLP:journals/corr/abs-2304-14108}, the models are evaluated on a total of 38 image classification and retrieval tasks to measure generalization to multiple different domains. The classification tasks are evaluated in a zero-shot manner. The aggregated results are displayed in \Cref{tab:results} and show the accuracy on the ImageNet classification task, the average performance on six datasets used to measure the robustness to distributions different from the ImageNet dataset \cite{DBLP:conf/nips/TaoriDSCRS20}, the average performance on the tasks from the Visual Task Adaptation Benchmark (VTAB) \cite{DBLP:journals/corr/abs-1910-04867}, the average performance on the three retrieval tasks (Flickr30k \cite{young-etal-2014-image}, MSCOCO \cite{DBLP:journals/corr/ChenFLVGDZ15} and WinoGAViL \cite{bitton2022winogavil}) and the overall average performance over all tasks. Keep in mind that higher values indicate better performance.

Note that the values for the baseline in \Cref{tab:results} differ slightly from the ones reported by Gadre et al. \cite{DBLP:journals/corr/abs-2304-14108}. This is due to the fact that the dataset used to train the models is not distributed by the authors and had to be retrieved using a provided script. Since some images and websites were not available anymore, the dataset is slightly different to the one Gadre et al. \cite{DBLP:journals/corr/abs-2304-14108} used to train the model.

The experiments were conducted on four NVidia 2080TI GPUs.

\subsection{Results}

\begin{table*}
	\begin{tabular}{l r r r r r}
		Training Data & ImageNet & ImageNet dist. shifts & VTAB & Retrieval & Average over 38 datasets \\
		CommonPool \cite{DBLP:journals/corr/abs-2304-14108} & 0.028 & 0.037 & 0.145 & 0.113 & 0.132 \\
		arXiv + PMC & 0.002 & 0.007 & 0.098 & 0.058 & 0.086 \\
		CommonPool + arXiv + PMC & 0.017 & 0.026 & 0.153 & 0.098 & \textbf{0.135} \\
	\end{tabular}
	\caption{Results of the small-scale CLIP model trained on different datasets.}
	\label{tab:results}
\end{table*}

Unsurprisingly, \Cref{tab:results} shows that the baseline model trained on CommonPool outperforms the model trained solely on arXiv and PMC, showing that the size and the domain coverage of the collected dataset are insufficient to train a general model alone. However, the model trained by extending CommonPool with arXiv and PMC subsets performs better on average than the baseline model. Together, these two observations suggest that including our collected dataset improves the performance of general CLIP model, even if the domain coverage is limited. However, it is clear that the difference in performance is rather small.

An important observation is that the performance gain is not uniform across tasks and domains. Although average performance is better, it is \emph{worse} on the ImageNet and Retrieval tasks. To investigate on which tasks and domains our dataset improves performance, \Cref{tab:tasks} lists the tasks on which the model trained on our dataset performs better than the baseline. In absolute terms, the most significant performance increase can be observed for tasks in the metastatic tissues domain, which might be explained by the large presence of this domain in the PMC subset. Still, performance also increases in multiple other domains, showing that the performance increase is not confined to a single domain overrepresented in our dataset.

\begin{table*}
	\begin{tabular}{lllllr}
		Task name & Type & Domain & Score & Baseline score & Reference \\
		FGVC Aircraft & classification & aircrafts & 0.0136 & 0.0072 & \cite{DBLP:journals/corr/MajiRKBV13} \\
		GTSRB & classification & traffic signs & 0.0801 & 0.0418 & \cite{Houben-IJCNN-2013} \\
		CLEVR Counts & counting & Blender-generated objects & 0.1465 & 0.1437 & \cite{DBLP:conf/cvpr/JohnsonHMFZG17} \\
		KITTI distance & distance prediction & vehicles & 0.3459 & 0.3150 & \cite{DBLP:conf/cvpr/GeigerLU12} \\
		PatchCamelyon & classification & metastatic tissues & 0.6004 & 0.4057 & \cite{DBLP:conf/miccai/VeelingLWCW18} \\
		SVHN & classification & house numbers & 0.1127 & 0.0852 & \cite{svhn} \\
		Camelyon17 & classification & metastatic tissue & 0.7120 & 0.3970 & \cite{DBLP:journals/tmi/BandiGMDBHBLPZL19} \\
	\end{tabular}
	\caption{Evaluation tasks on which the model trained on the arXiv and PMC datasets performs better than the baseline model.}
	\label{tab:tasks}
\end{table*}

\section{Conlusion}

In conclusion this work has shown that using high-quality scientific image-text pairs in addition to existing large-scale image-text pairs crawled from the web improves the performance of CLIP models. However, the performance improvement is rather modest and is not uniform across evaluation tasks.

\subsection{Limitations}

The experiments in this paper were only performed on the small scale model and data of Gadre et al. \cite{DBLP:journals/corr/abs-2304-14108}. Experiments by them show that insights from the small scale do not necessarily translate to larger scales. Furthermore, not all possible data could be extracted due to resource limitations.

Additionally, the collected datasets are not deduplicated, neither within the subset nor against the other subset.

\subsection{Future Work}

The insights derived in this paper lead to the obvious next step to extract all available data from arXiv and PMC and train larger scale models using this data. The results of such an experiment would provide an indication of whether the scheme described in this paper is worth extending with other data sources.

The dataset collected in this paper can easily be extended by a magnitude or more, thereby preserving the dataset proportions at larger scales without necessitating upsampling. Consider the PMC subset, which only consists of data from 10 out of 255 available directories. Preliminary exploration using a script shows that in total 18065333 figures could be extracted. Similarly, the arXiv data can be extended by considering papers from 2021-01-01 onwards. As the source files of this time period make up 1.6TB - which is roughly similar to the source files considered in this paper - we predict that the dataset size can be more than doubled. Moreover, this trend leads to believe that the amount of papers hosted on arXiv (and thus the available data) will only continue to increase exponentially.

Furthermore, improvements to the data collection pipeline are possible. While the PMC data is available in XML format and can be easily parsed, the arXiv data is available in \TeX format, which ideally needs to be compiled to be accurately parsed. A promising avenue is to use LaTeXML to convert the \TeX files into XML files. This solves some drawbacks of using TexSoup \cite{TexSoup}, as it supports more \LaTeX features (most importantly the \verb.\newcommand. command). Additionally, a better way to handle references and citations in captions (instead of just replacing them with \verb|<ref>| and \verb|<cit.>|) could be found.

Moreover, the text corresponding to an image could be extended by using every sentence containing a reference to the image figure in the paper.

Since the subsets are dominated by complicated graphs and plots, another possibility is to remove all or a large portion of this kind of data, leaving mostly natural images. However, this is of dubious benefit, since there are bound to be visual tasks relating to graphs and plots specifically.

\section{Acknowledgments}
Thanks to TU Wien for providing the computing resources and to the AWS Public Sector Cloud Credit for Research program for covering the cost of downloading data from the arXiv S3 bucket.

{\small
\bibliographystyle{ieee_fullname}
\bibliography{bib}
}

\end{document}